\documentclass[preprint,review,12pt]{elsarticle}

\usepackage{amssymb}
\usepackage[]{graphicx}
\usepackage{booktabs}
\usepackage{amsmath}
\usepackage{subfigure}

\usepackage[ruled,linesnumbered, vlined]{algorithm2e}

\journal{Signal Processing}

\begin{document}

\begin{frontmatter}



\title{Finding More Relevance: Propagating Similarity on Markov Random Field for Image Retrieval}


\author[1]{Peng Lu}
\author[2]{Xujun Peng}
\author[3]{Xinshan Zhu}
\author[1]{Xiaojie Wang}

\address[1]{Beijing University of Posts and Telecommunications, Beijing, 100876, China}
\address[2]{Raytheon BBN Technologies, Cambridge, MA, USA}
\address[3]{Tianjing University, Tianjin, 300072, China}

\begin{abstract}
To effectively retrieve objects from large corpus with high accuracy is a challenge task. In this paper, we propose a method that propagates visual feature level similarities on a Markov random field (MRF) to obtain a high level correspondence in image space for image pairs. The proposed correspondence between image pair reflects not only the similarity of low-level visual features but also the relations built through other images in the database and it can be easily integrated into the existing bag-of-visual-words(BoW) based systems to reduce the missing rate. We evaluate our method on the standard Oxford-5K, Oxford-105K and Paris-6K dataset. The experiment results show that the proposed method significantly improves the retrieval accuracy on three datasets and exceeds the current state-of-the-art retrieval performance.

\end{abstract}


\end{frontmatter}


\section{Introduction}
In this paper, we address the problem of image retrieval where the goal is to retrieve all images containing a particular query object that is outlined by users in a large scale and unordered image database. Although there are many literatures explored several approaches to solve this problem, to retrieve images according to query object in real time with high accuracy is still an extremely challenge task. Amongst the existing image retrieval methods, most early work in this field are originated from the BoW scheme, which is inspired by the method from text-retrieval systems \cite{Sivic03}. Despite of its simplicity and efficiency, BoW based image retrieval methods are heavily relied on low-level information whose discriminative capability is degraded quickly in high dimensional spaces \cite{QinGBQG11}. To overcome this problem, Philbin \cite{Philbin10a} suggested an approximate nearest neighbor based k-means approach to build visual vocabulary, which boosted the retrieval accuracy considerably.

In order to avoid the false rejection problem introduced from the vocabulary quantization, Philbin described a soft-assignment method in \cite{Philbin10a} which can include features lost in the quantization stage. In \cite{Arandjelovic12}, Arandjelovi\'c proposed a square root (Hellinger) kernel based approach to measure the distance between SIFT descriptors, which outperformed the standard Euclidean distance measure for image retrieval. Similarly, J\'egou \cite{JDS10a} presented a Hamming embedding scheme to represent descriptors more precisely where weak geometric consistency constraints were applied. \par

Instead of focusing on improvements on the vocabulary quantization, other researchers explored methods on the descriptor representation. In \cite{Perdoch-CVPR-2009}, Per{\v d}och suggested an improved feature detector where discretized local geometry representation was learnt. By exploring different configuration of normalization, dimension reduction and dynamic range reduction, Simon proposed a better visual descriptor other than SIFT for image retrieval with high discriminative power \cite{DAISY}.

To deal with feature detection drop-out, Chum\cite{Chum07b}\cite{ChumMPM11} adopted a set of query expansion methods to enrich the query model by using spatially verified features. In \cite{Arandjelovic12}, Arandjelovi\'c proposed a discriminative query expansion approach, where only the weights of positive words were learnt.\par

 Inspired by but unlike the method described in \cite{QinGBQG11} and \cite{WangWZTGL12}, 
 in this study we propose a concept of comprehensive relevance (CR) that incorporates the low-level feature similarity into a high level relevance measure through Markov random field (MRF). In this framework, the correspondence between image pair is computed not only relied on their low-level visual features, but also dependent on the relations through other images in the database. The proposed approach can be easily integrated into the existing BoW based systems to reduce the missing rate.

We organize the rest of the paper as follows. Section \ref{motivation} introduces the concept of comprehensive relevance, followed by the implementation details of CR in section \ref{implementation}.  The experimental setup and results are provided in section \ref{experiment}. Section \ref{conlusion} concludes the paper.

\section{Motivation \& Problem Formulation\label{motivation}}
Given a query image $x_q$ and an image set $X$ which is represented by an undirected graph $G(V,E)$, a random field with respect to $G(V,E)$ can be formed where random variables $x_v \in X$ are indexed by $V$ and connected by $E$ which measure the similarity between image pairs in dataset. Then, the object retrieval task can be modeled as a random field optimization problem by maximizing the posterior:
\begin{equation} \label{eq:random_field}
    \hat{X} = \arg\max_{X}P(X \mid x_q) = \arg\max_{X} \prod_{v=1}^{|V|}P(x_v \mid X_{V-v}, x_q)
\end{equation}
where $V-v$ is the set of all vertices in graph $G$ except vertex $v$. This equation shows that each image's type (object/non-object) is dependent on its relevance to the query image and all other images in the dataset.\par

By taking Bayes rule and Markov property, Eq. (\ref{eq:random_field}) can be rewritten as:
\begin{align}
    \nonumber \hat{X} & = \arg\max_X \prod_{v=1}^{|V|} P(x_v \mid X_{\Gamma(v)}, x_q)  = \arg\max_X \prod_{v=1}^{|V|} P(x_q \mid x_v, X_{\Gamma(v)}) P(x_v \mid X_{\Gamma(v)})\\
                      & = \arg\max_X \sum_{v=1}^{|V|} \log P(x_q \mid x_v, X_{\Gamma(v)}) + \log P(x_v \mid X_{\Gamma(v)})
\end{align}
where prior $P(x_v \mid X_{\Gamma(v)})$ means the property of $x_v$ is conditioned by its neighbors $X_{\Gamma(v)}$ and likelihood $P(x_q \mid x_v, X_{\Gamma(v)})$ describes the relation of query image $x_q$ and image $x_v$. \par

In our work, we approximate $\log P(x_q \mid m_v, X_{\Gamma(v)})$ by using a similarity function $d_v(q)$ which is called \textit{direct relevance} between $x_v$ and $x_q$. To compute $\log P(x_v \mid X_{\Gamma(v)})$, we apply a similarity function $\tau_v(q)$ by using belief propagation technique to obtain \textit{indirect relevance} between $x_v$ and $x_q$ from $x_v$'s neighbors. The summation of direct relevance $d_v(q)$ and indirect relevance $\tau_v(q)$ is called \textit{comprehensive relevance} $s_v(q)$ in this paper. More details of function $d_v(q)$ and $\tau_v(q)$ are described in the following section. \par

Prior to the computation of $d_v(q)$ and $\tau_v(q)$ between images, a graph $G(V,E)$ corresponding to the image corpus is created where each edge $E(u, v)$ indicates that there is a match between corresponding image pair $x_u$ and $x_v$ which is measured by a positive weight $m_{u,v}$. In our work, we use inlier correspondences to measure the match $m_{u,v}$ between two images $x_u$ and $x_v$ in corpus:
\begin{equation} \label{eq:match}
    m_{u,v} =
    \begin{cases}
        \frac{c_{uv}^2}{\sigma^2+c_{uv}^2}, \hspace{2mm} c_{uv} \geq \theta \\
        0, \hspace{10mm} c_{uv} < \theta
    \end{cases}
\end{equation}
where $c_{uv}$ is the inliers number between two images, $\theta$ is a predefined threshold and $\sigma$ is scale parameter. 

\begin{figure}[h]
    \centering
    \includegraphics[width=300pt]{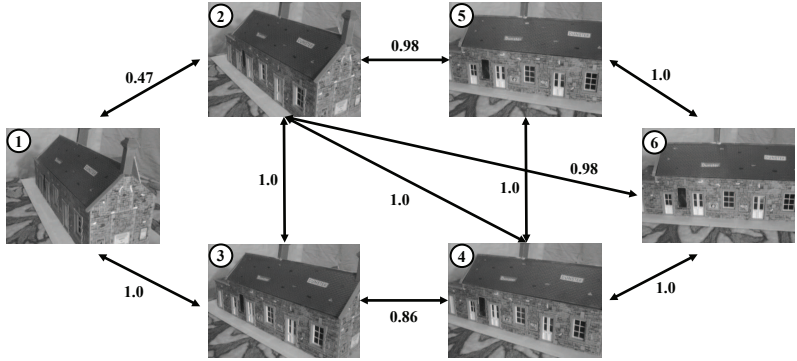}
    \caption{Examples of the same object with different viewpoints. \label{fig:Graph}}
\end{figure}

According to Eq. (\ref{eq:match}), the match score of every image pair in dataset can be obtained to create a graph. For example, as shown in Fig. \ref{fig:Graph}, a graph is generated for the sample images which contain the same target with different viewpoints. Based on the created graph, there exists pathes direct/indirect connecting query object 1 with every other images, even though there is not edges direct connect them due to different viewpoint (such as object 6). Intuitively, in the concept of CR, the similarity between two images is calculated according to the direct relevance which measures the weight of direct path and the indirect relevance which is computed based on the number of edges on the indirect path that connects two images and the weights associated with them. Therefore, CR can take the advantage of BoW based method with lower false rejection rate.

\section{Implementation of Comprehensive Relevance\label{implementation}}
\subsection{Comprehensive Relevance Update Criteria}
As described in the previous section, given a graph $G(V,E)$ of a database and a query image $x_q$ that does not belonged to $V$, we define vectors $\boldsymbol{d}, \boldsymbol\tau, \boldsymbol{s} \in \mathcal{R}^{|V| \times 1}$ whose elements $d_{v}(q), \tau_{v}(q), s_{v}(q)$ represent the direct relevance, indirect relevance and comprehensive relevance for $x_v$ and $x_q$, accordingly. \par

To the direct relevance $d_v(q)$, the feature level similarity defined in Eq. (\ref{eq:match}) can be used directly to estimate the match degree between two images:
\begin{equation} \label{direct_relevance}
    d_v(q) = m_{v,q}
\end{equation}

The idea behind indirect relevance $\tau_v(q)$ is to find the relationship between two images through other images, which are relied on both the number of edges and the entire weights between them. Intuitively, the similarity score is higher between two images if the corresponding vertices in the graph $G(V,E)$ have more pathes containing shorter edges with larger weights. In our work, we implemented a belief propagation based approach to pass the similarity message between vertex pairs through entire network of graph $G$ and used them to calculate the indirect relevance. \par

During the message propagation, each vertex in $G$ propagates its belief to its direct neighbors via weighted edges, which can be described as:
\begin{equation} \label{eq:belief1}
    \boldsymbol\tau^{k+1} = A \boldsymbol\tau^k
\end{equation}
where superscript $k$ is the iteration, $A$ is a weighted adjacency matrix of graph $G$ whose element $a_{ij}$ is normalized weight: $a_{ij}=\frac{m_{ij}}{\sum_{n=1}^{|V|} m_{in}}$, where $i,j \in |V|$.

In the real large corpus, multiple objects may be contained within the same images that causes the path between two different objects. To overcome this problem, we introduce a decay factor $\alpha$ during belief propagation to penalize long propagation distance between any two vertices and a residual factor $\beta \boldsymbol{d}$ to compensate the initial similarity belief. By taking $\beta=1-\alpha$, Eq. (\ref{eq:belief1}) can be re-written as:
\begin{equation} \label{eq:indirect_relevance}
    \boldsymbol\tau^{k+1} = \alpha A \boldsymbol\tau^k + (1-\alpha) \boldsymbol{d}
\end{equation}

The resulting advantage of Eq. (\ref{eq:indirect_relevance}) is that it has the same form as biased PageRank algorithm \cite{ilprints386} which has been proved its efficiency in large dataset.

Prior to belief propagation, $\boldsymbol\tau$ is initialized by $\boldsymbol\tau = \boldsymbol{d}$ and updated according to Eq. (\ref{eq:indirect_relevance}) by $N$ times, where $N$ is the propagation distance which determines the length that a message can be propagated in the graph $G(V,E)$. 

Based on Eq. (\ref{direct_relevance}) and Eq. (\ref{eq:indirect_relevance}), the comprehensive relevance is defined as:
\begin{equation} \label{comprehensive_relevance}
    \boldsymbol{s} = \gamma \boldsymbol{d} + (1-\gamma) \boldsymbol\tau
\end{equation}
where $\gamma$ is a parameter which takes value in the range of $[0,1]$ and controls the influences from direct relevance and indirect relevance.

The overall approach of CR computation is summarized in Algorithm \ref{algo:cr}.

\begin{algorithm} \label{algo:cr}
  \caption{Relevance computation algorithm}
    \SetKwInOut{Input}{input}
    \SetKwInOut{Output}{output}
    \SetKwFunction{ComputeInliers}{ComputeInliers}

    \Input{$A$: weighted adjacency matrix \\
           $v_q$: query image \\
           $\alpha$: decay factor \\
           $\gamma$: control factor \\
           $N$: iterations}
    \Output{$\textbf{s}$: Comprehensive relevance scores of vertex set $V$}

    \BlankLine
    \For{$i \leftarrow 1$ \KwTo $|V|$} {
        $c_{iq} \leftarrow$ \ComputeInliers{$v_i$, $v_q$}\;
        \lIf{$c_{iq} \geq \theta$} {$d_i=c_{iq}^2/(\sigma^2+c_{iq}^2)$}
        \lElse{$d_i=0$}
    }

    $\boldsymbol{d} \leftarrow \boldsymbol{d}/\sum_{i=1}^{|V|}d_i$\;
    $\boldsymbol\tau^0 \leftarrow \boldsymbol{d}$\;
    \For{$n \leftarrow 1$ \KwTo $N$} {
        $\boldsymbol\tau^n = \alpha A \boldsymbol\tau^{n-1} + (1-\alpha)\boldsymbol{d}$ \;
    }
    $\boldsymbol{s} \leftarrow \gamma \boldsymbol{d}+(1-\gamma)\boldsymbol\tau^N$ \;
\end{algorithm}

\subsection{Optimal Subgraph Construction \label{subgraph}}
It can be seen from Eq. (\ref{comprehensive_relevance}) that the computation complexity of $\boldsymbol{s}$ is dependent on the size of adjacent matrix $A$, which is determined by the order of $G(V,E)$. In the real applications, most images in the dataset are not related with query image. Thus, in order to reduce the computation complexity, 
we designed a scheme to extract an optimal subgraph $G^*(V^*, E^*)$ from graph $G(V,E)$ whose order $|V^*|$ is relatively small enough and $V^*$ contains most relevant images. \par

Given a graph $G$ of a dataset, a small set of vertices which have higher correspondence to the query image can be initially obtained from $G$. We call these vertices as \textit{Root Vertex Set} $R$. To find an optimal subgraph $G^*(V^*, E^*)$ from $G(V,E)$, we can expand the size of $R$ to include more vertices that are similar to query image. Thus, to each vertex $v \in R$, its neighbors $\Gamma(v)$ are merged into the optimal subgraph $G^*$. This procedure can be taken $M$ times to ensure most vertices that contain the same object are included. Algorithm \ref{algo:optimal_graph} briefly describes the steps of optimal subgraph extraction.

\begin{algorithm} \label{algo:optimal_graph}
  \caption{Optimal Subgraph Algorithm}
    \SetKwInOut{Input}{input}
    \SetKwInOut{Output}{output}

    \Input{$G(V,E)$: matching graph of data set \\
           $R$: Root vertex set \\
           $M$: optimal searching depth}
    \Output{$G^*(V^*, E^*)$: Optimal subgraph of $G(V,E)$}

    \BlankLine
    $V^* \leftarrow R$, $V_t \leftarrow R$   \tcc*{$V_t$ is a temporary variable.}
    \For{$i \leftarrow 1$ \KwTo $M$} {
        \For{$v_j \in V_t$} {
            $V^* \leftarrow V^*\cup\Gamma(v_j)$ \;
        }
        $V_t \leftarrow V^*$ \;
    }
    $G^* := G[V^*]$ \;
 \end{algorithm}

\section{Experimental Results \label{experiment}}
To evaluate the proposed algorithm, we performed experiments on three standard benchmark datasets: \textbf{Oxford-5K} dataset \cite{Philbin10a}, \textbf{Oxford-105K} dataset \cite{Philbin10a} and \textbf{Paris-6K} dataset \cite{Philbin10a}. To achieve the high image retrieval accuracy with low computation complexity, we implemented a benchmark image retrieval system as described in \cite{Perdoch-CVPR-2009, Philbin10a} where similarity between image pairs in the dataset were calculated using cosine of angle and the top ranked image pairs were re-ranked using inlier correspondences. On the top of this benchmark system, a graph $G$ was created initially. Then, a root vertex set $R$ was determined according to top ranked image list which was expanded to an optimal subgraph $G^*(V^*, E^*)$ as described in section \ref{subgraph}. Based on $G^*(V^*,E^*)$, comprehensive relevance can be calculated and the objects in the dataset can be retrieved efficiently. \par

\subsection{Analysis of Parameters}
\begin{figure} [h]
    \centering
    \subfigure[Recall analysis]{\includegraphics[width=2.6in]{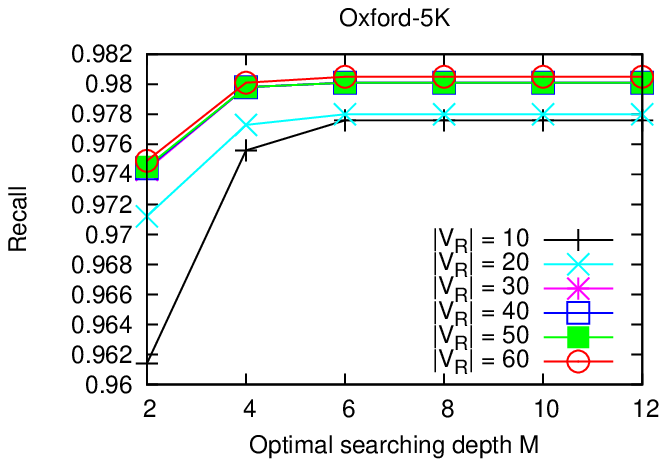} \label{oxford5k_recall}}
    \subfigure[Order of $G^*$]{\includegraphics[width=2.6in]{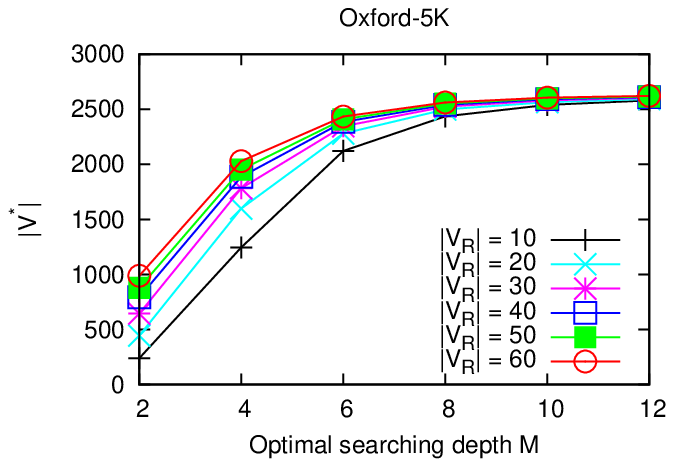} \label{oxford5k_size}}
    \caption{Effects of $|V_R|$ and $M$ on Oxford-5K set.}  \label{oxford5k_setting1}
\end{figure}

\begin{figure} [h]
    \centering
    \subfigure[Recall analysis]{\includegraphics[width=2.6in]{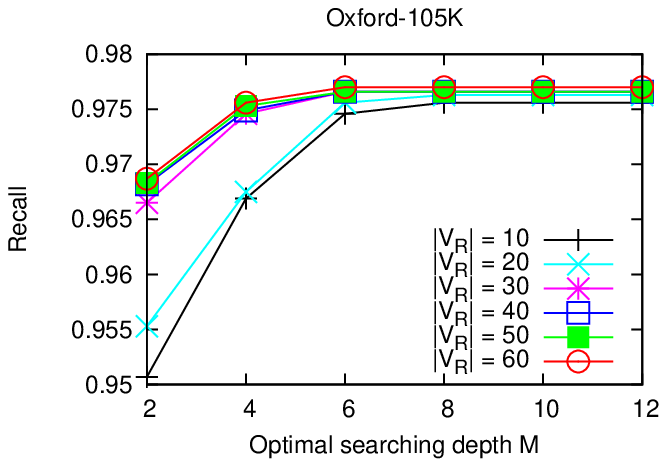} \label{oxford105k_recall}}
    \subfigure[Order of $G^*$]{\includegraphics[width=2.6in]{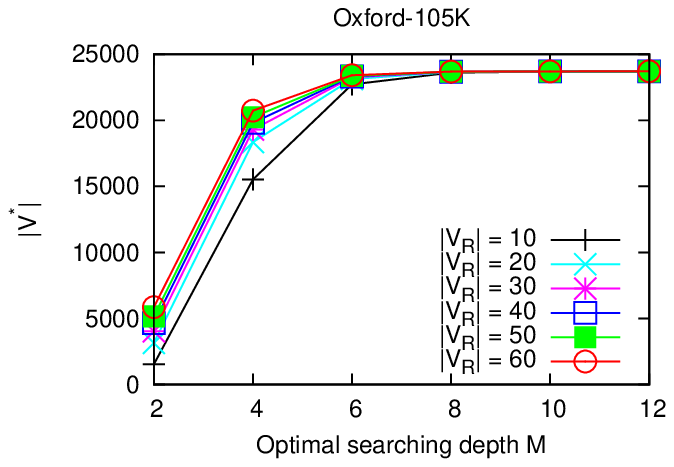} \label{oxford105k_size}}
    \caption{Effects of $|V_R|$ and $M$ on Oxford-105K set.} \label{oxford105k_setting1}
\end{figure}

\begin{figure} [h]
    \centering
    \subfigure[Recall analysis]{\includegraphics[width=2.6in]{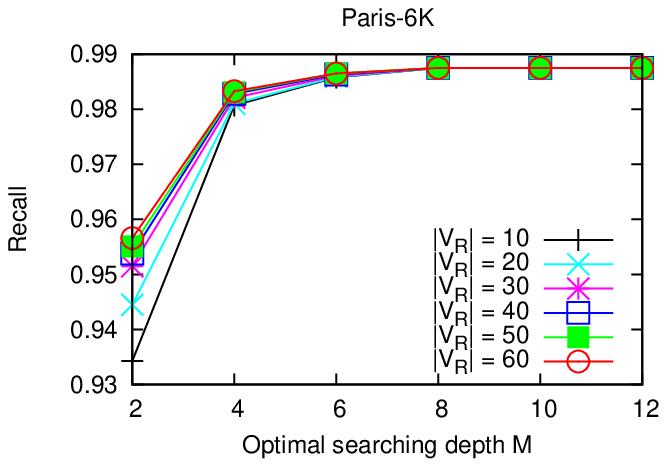} \label{paris6k_recall}}
    \subfigure[Order of $G^*$]{\includegraphics[width=2.6in]{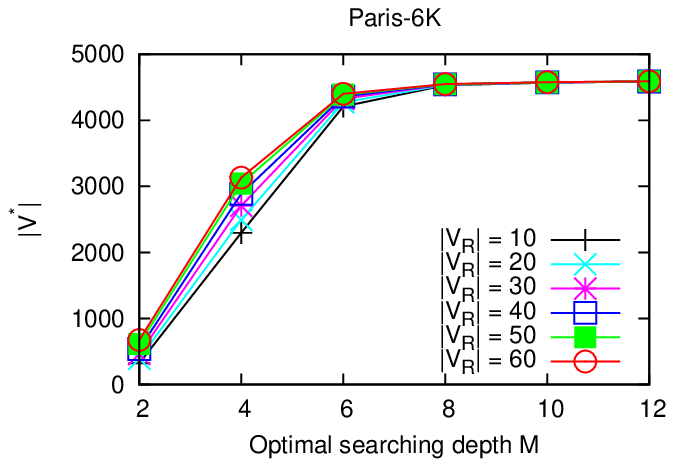} \label{paris6k_size}}
    \caption{Effects of $|V_R|$ and $M$ on Paris-6K set.} \label{paris6k_setting1}
\end{figure}

In our experiments, we first analyzed the effects of parameter root vertex set size $|V_R|$ and optimal searching depth $M$ on the performance. In Fig. \ref{oxford5k_recall}, we illustrated the recall on Oxford-5K set with different settings of $|V_R|$ and $M$. The corresponding optimal graph's order $|V^*|$ with the same settings was shown in Fig. \ref{oxford5k_size}. The same experiments were carried out on Oxford-105K set and Paris-6K set as shown in Fig. \ref{oxford105k_setting1} and Fig. \ref{paris6k_setting1}, respectively.

As discussed in previous section and can be seen from Fig. \ref{oxford5k_setting1}, \ref{oxford105k_setting1} and \ref{paris6k_setting1}, root set size $|V_R|$ and optimal searching depth $M$ determine the order of optimal graph $G^*$ and the retrieval accuracy, where as the $|V_R|$ and $M$ increase, both the accuracy and the computation complexity are increased consequently. Generally, $M$ tends to be inversely proportional to the size of dataset $|V|$ to ensure both accuracy and efficiency.  Based on the observation from above figures, we set $M=3$ to Oxford-5K and Paris-6K sets, $M=2$ to Oxford-105K set, and $|V_R|=30$ to all three datasets in our experiments. \par

\begin{figure} [h]
    \centering
    \subfigure[]{\includegraphics[width=2.6in]{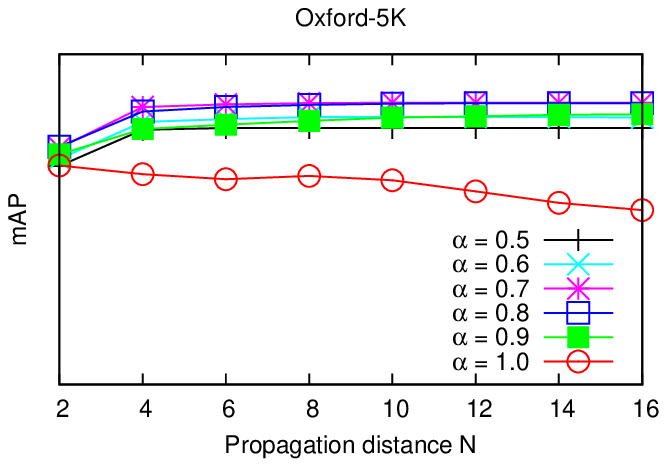} \label{oxford5k_M_R}}
    \subfigure[]{\includegraphics[width=2.6in]{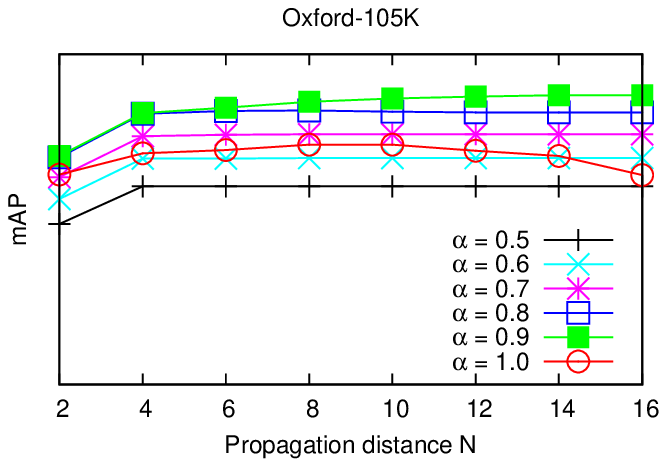} \label{oxford105k_M_R}}
    \subfigure[]{\includegraphics[width=2.6in]{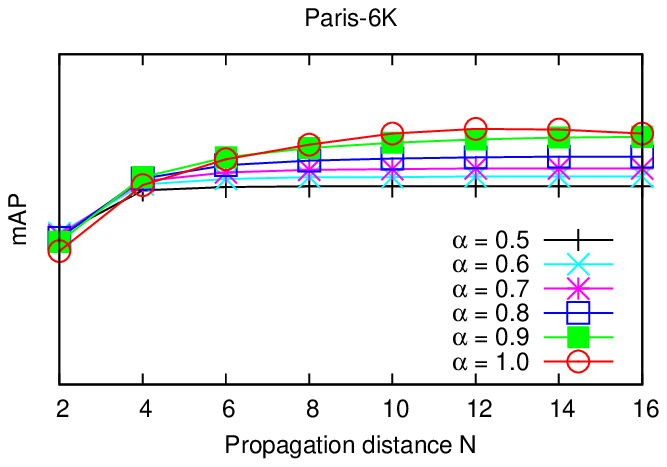} \label{paris6k_M_R}}
    \caption{mAP vs. propagation distance $N$ and decay factor $\alpha$. (a) On Oxford-5K set. (b) On Oxford-105K set. (c) On Paris-6K set.} \label{system_setting2}
\end{figure}

Looking deeper into the proposed method, we measured mean average precision (mAP) on three datasets using different propagation distance $N$  and decay factor $\alpha$, as shown in Fig. \ref{system_setting2}. From this figure, we can see that normally a higher mAP is achieved by increasing the propagation distance $N$ during CR update with $\alpha<1.0$. In our experiment, we set $N=10$ as a tradeoff between the accuracy and efficiency.  To the parameter $\alpha$, it can be observed that the performance of our system was degraded with very large or small $\alpha$ (typically $\alpha>0.9$ or $\alpha \le 0.5$) on Oxford-5K and Oxford-105K sets. By analyzing the content of each dataset, we found that better mAP can be obtained with a moderate $\alpha$ (such as $\alpha=0.6$) if the hub number of graph $G$ is big \cite{Grauman2008226}, which indicates large number of occurrences of multiple objects contained in the same image. To the dataset that does not have hub set, a larger $\alpha$ (such as $\alpha=1.0$) can be used to encourage extensive message propagation to find more similar objects as the query image. \par

\subsection{Comparison with the state-of-the-art Approaches}
\begin{figure} [h]
    \centering
    \subfigure[]{\includegraphics[width=2.6in]{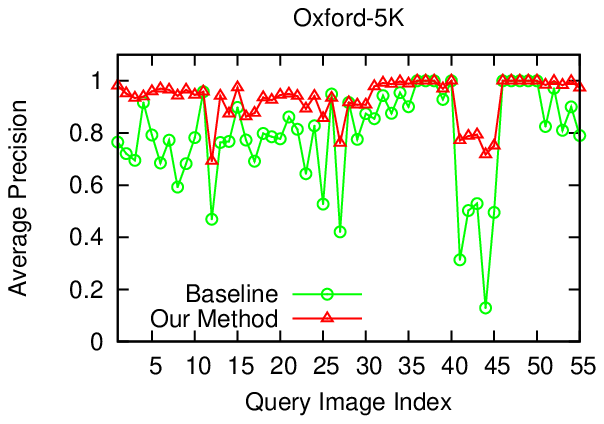} \label{oxford5k_query}}
    \subfigure[]{\includegraphics[width=2.6in]{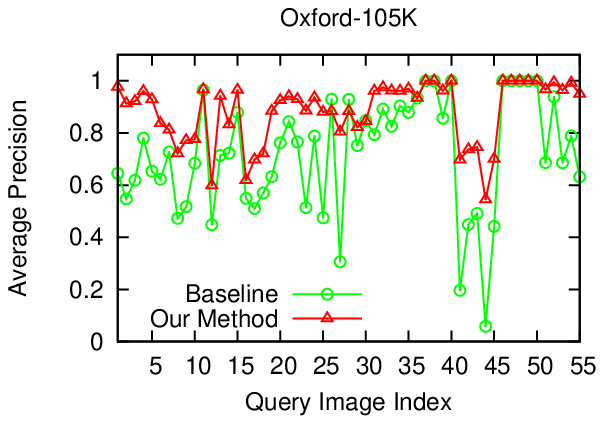} \label{oxford105k_query}}
    \subfigure[]{\includegraphics[width=2.6in]{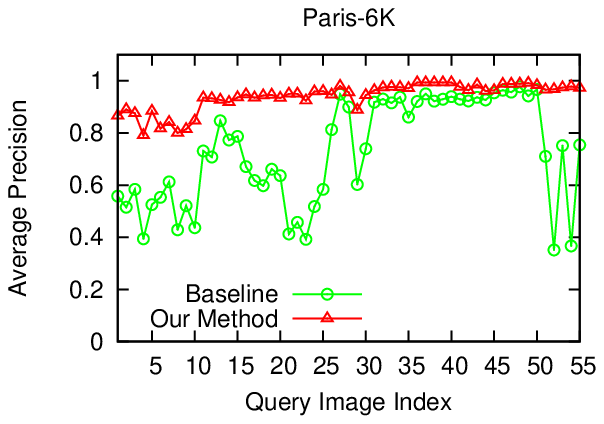} \label{paris6k_query}}
    \caption{Comparison of average precision with benchmark and the proposed system.} \label{ap_comparison1}
\end{figure}

In this subsection, we firstly compared the proposed image retrieval method with the benchmark system on three datasets. In Fig. \ref{ap_comparison1}, the average precisions (AP) of the benchmark system and the proposed system on each dataset were shown, from which we can see the AP of 85.6\% queries has improved and 11.5\% remained unchange using the proposed method. Within those queries without improvement, 89.7\% of them already have AP of 1, which indicates the proposed method improved precision on 95.8\% queries.

Furthermore, we compared the proposed method with the baseline and the state-of-the-art approaches. As shown in Table \ref{tab:exp:the-state-of-art}, the mAPs of the proposed method were greatly improved by 18.1\%, 28.8\% and 22.8\% on Oxford-5K, Paris-6K and Oxford-105K set relatively compared to benchmark system. For the set of Oxford-105K, our method miss the state-of-the-art only by 0.007 of mAP. For Oxford-5K and Paris-6K set, the proposed method competes with the state-of-the-art approaches, which demonstrates a significant improvement achieved by our method. \par

\begin{table}[h]
\centering
\small
\caption{mAP comparisons on different datasets.}
\begin{tabular}{cccc}
\hline
Dataset & Oxford-5K  & Oxford-105K & Paris-6K \\
\hline
Baseline & 0.789  & 0.72 & 0.73 \\
Proposed method & \textbf{0.932}  & 0.884  & \textbf{0.94}\\
 Philbin $et$ $al$. \cite{Philbin10a} & 0.825 &   0.719 & - \\
Per{\v d}och $et$ $al$. \cite{Perdoch-CVPR-2009}     & 0.916 &  0.885  & -  \\
Mikulik $et$ $al$. \cite{Mikulik-ECCV-2010}    &  0.849 &	0.795 &	0.824  \\
Qin $et$ $al$. \cite{QinGBQG11}    &  0.814  &	0.767 &	0.803  \\
Shen $et$ $al$. \cite{ShenLBAW12} & 0.884  & 0.864&  0.911   \\
Arandjelovi\'c $et$ $al$. \cite{Arandjelovic12} & 0.929 & \textbf{0.891}&  0.910   \\
\hline
\end{tabular}
\label{tab:exp:the-state-of-art}
\end{table}

\section{Conclusions \label{conlusion}}
Unlike previous image retrieval methods that only use low-level similarity in visual feature space to rank images, in this paper, we achieve high retrieval accuracy by propagating feature level similarity on a MRF to find the correspondence between image pairs via other images, with an efficient optimal subgraph construction method. Experiments show that the proposed method takes advantages from existing retrieval systems and significantly outperforms the state-of-the-art approaches for both accuracy and efficiency. 









\end{document}